%% file: main.tex
\documentclass[10pt]{article} 

\usepackage[accepted]{rlj} 

\usepackage{amssymb}            
\usepackage{mathtools}          
\usepackage{mathrsfs}           
\usepackage{graphicx}           
\usepackage{subcaption}         
\usepackage[space]{grffile}     
\usepackage{url}                
\usepackage{lipsum}             
\usepackage{listings}
\usepackage{booktabs}
\usepackage{placeins}
\usepackage{amsmath,amssymb,amsfonts}
\usepackage{algorithm}
\usepackage{algpseudocode}
\usepackage{bbm}
\newtheorem{theorem}{Theorem}

\title{Grow-Prune-Freeze Networks: Adaptive \& Continual Learning Technique for Olfactory Navigation}

\setrunningtitle{Grow-Prune-Freeze Networks}

\author{Kordel K. France\textsuperscript{1}, Ovidiu Daescu\textsuperscript{1}}

\emails{kordel.france@utdallas.edu, ovidiu.daescu@utdallas.edu}

\affiliations{
$^{1}$\textbf{Department of Computer Science, University of Texas at Dallas, USA}\\
}

\contribution{
    A framework for adaptive and continual reinforcement learning agents operating in partially observable, non-stationary environments and grounded in random matrix theory (RMT).
    }
    {
    We show that foundational pruning mechanisms from \cite{lecun1989optimalbraindamage} can be used to improve policy gradient methods whilst simultaneously growing a multi-layer perceptron and freezing earlier layers to hedge against catastrophic forgetting.
    }

\contribution{
    RMT analysis showing spectral eigenvalue invariance for indefinitely adapting models, providing a mechanism for resource-efficient learning in robotics that enables an agent to allocate representational capacity where environmental signals are complex or changing. 
    }
    {
    Work by \cite{Pennington2017} and \cite{Martin2021RMT} show that spectral invariance among eigenvalues of weight matrices can be preserved among neural networks with a single hidden layer. We extend their methods to show that their assumptions support neural networks with multiple hidden layers.
    }

\contribution{
    Tuning criterion for introduced parameters to stabilize continual learning with GPFs and an Expected SARSA policy.
    }
    {
    Prior work by \cite{france2026_ghosts_mdpi} highlights the potential of Expected SARSA in olfactory navigation and we show that their findings extrapolate well to our GPF framework.
    }
\contribution{
    Empirical demonstration that GPFs achieve 98\% source-finding success on a partially observable olfactory navigation task, generalizing to 94\% on held-out trials across diverse random initializations and notable performance on non-olfactory tasks.
    }
    {
    GPFs were specifically designed to solve a problem in partially observable olfactory tasks. However, we demonstrate our methods over multiple tasks to build support for generalization of GPFs in other environments, but do not claim state-of-the-art performance across all benchmarks.
    }

\keywords{olfaction, policy gradient, temporal difference learning, robotics, world models} 

\summary{Training data for olfaction is scattered through disparate, non-standardized datasets that limit the ability to build representative world models.
Olfactory navigation is a highly dynamic and non-stationary task that benefits from real-time continual learning.
We introduce an adaptive framework called \textit{Grow-Prune-Freeze (GPF) networks} that enable an agent to continually learn through growing, pruning, and freezing early layers of its policy in response to world complexity. 
Grounding GPFs in non-linear random matrix theory, we show that the work of \cite{Pennington2017} can be extended from single hidden layers to $n$-layer continual-learning models, and that eigenvalue composition of network weights is preserved as successive layers are added.
We show that GPFs based on Expected SARSA achieve a 94\% success rate on turbulent plume navigation---a partially observable, non-stationary task representative of the ``big world'' challenges that motivate adaptive learning in robotics---and provide supporting methodology for applying GPFs in other world models.
Further experiments amount evidence that GPFs may generalize well to other machine learning tasks such as reinforcement learning in Atari, image classification, and autoregressive language models.
We open source all code and data to encourage improvements on and more research in olfactory robotics.
}

\begin{document}

\makeCover  
\maketitle  

\begin{abstract}
\input{src/_abstract}
\end{abstract}

\input{src/_intro}
\input{src/_related_work}
\input{src/_methods_a}
\input{src/_experiments_a}
\input{src/_results}
\input{src/_future_work}
\input{src/_conclusion}

\input{src/_impact_statement}






\subsubsection*{Acknowledgments}
\label{sec:ack}
We would like to acknowledge the efforts of Scentience, Inc. in providing valuable hardware and sensor models that made the olfactory simulations possible. We would also like to thank John Machado and Rohith Peddi for their feedback on methodology and experiments.


\bibliography{main}
\bibliographystyle{rlj}

\beginSupplementaryMaterials
\setcounter{section}{0} 
\setcounter{subsection}{0} 
\renewcommand{\thesection}{\Alph{section}} 
\renewcommand{\thesubsection}{\thesection.\arabic{subsection}} 
\input{src/supplementary/_framework_table}
\input{src/supplementary/_pseudocode}
\input{src/supplementary/_rmt_background}
\input{src/supplementary/_homogeneous_layers}
\input{src/_methods_b}
\input{src/supplementary/_methods_c}
\input{src/supplementary/_simulator_details}
\newpage
\input{src/supplementary/_experiment_hyperparams}
\newpage
\input{src/supplementary/_results_table}
\newpage
\input{src/supplementary/_experiments_b}

%
%

\end{document}

%% file: src/_abstract.tex
\vspace{-1mm}
Training data for olfaction is scattered through disparate, non-standardized datasets that limit the ability to build representative world models.
Olfactory navigation is a highly dynamic and non-stationary task that benefits from real-time continual learning.
We introduce an adaptive framework called \textit{Grow-Prune-Freeze (GPF) networks} that enable an agent to continually learn through growing, pruning, and freezing early layers of its policy in response to world complexity. 
Grounding GPFs in non-linear random matrix theory, we show that the work of \cite{Pennington2017} can be extended from single hidden layers to $n$-layer continual-learning models, and that eigenvalue composition of network weights is preserved as successive layers are added.
We show that GPFs based on Expected SARSA achieve a 94\% success rate on turbulent plume navigation---a partially observable, non-stationary task representative of the ``big world'' challenges that motivate adaptive learning in robotics---and provide supporting methodology for applying GPFs in other world models.
Further experiments amount evidence that GPFs may generalize well to other machine learning tasks such as reinforcement learning in Atari, image classification, and autoregressive language models.
We open source all code and data to encourage improvements on and more research in olfactory robotics.


\vspace{-1mm}

%% file: src/_intro.tex
\section{Introduction}
\label{sec:intro}

\vspace{-1mm}
\begin{figure}[!b]
  \centering
  \includegraphics[width=\linewidth]{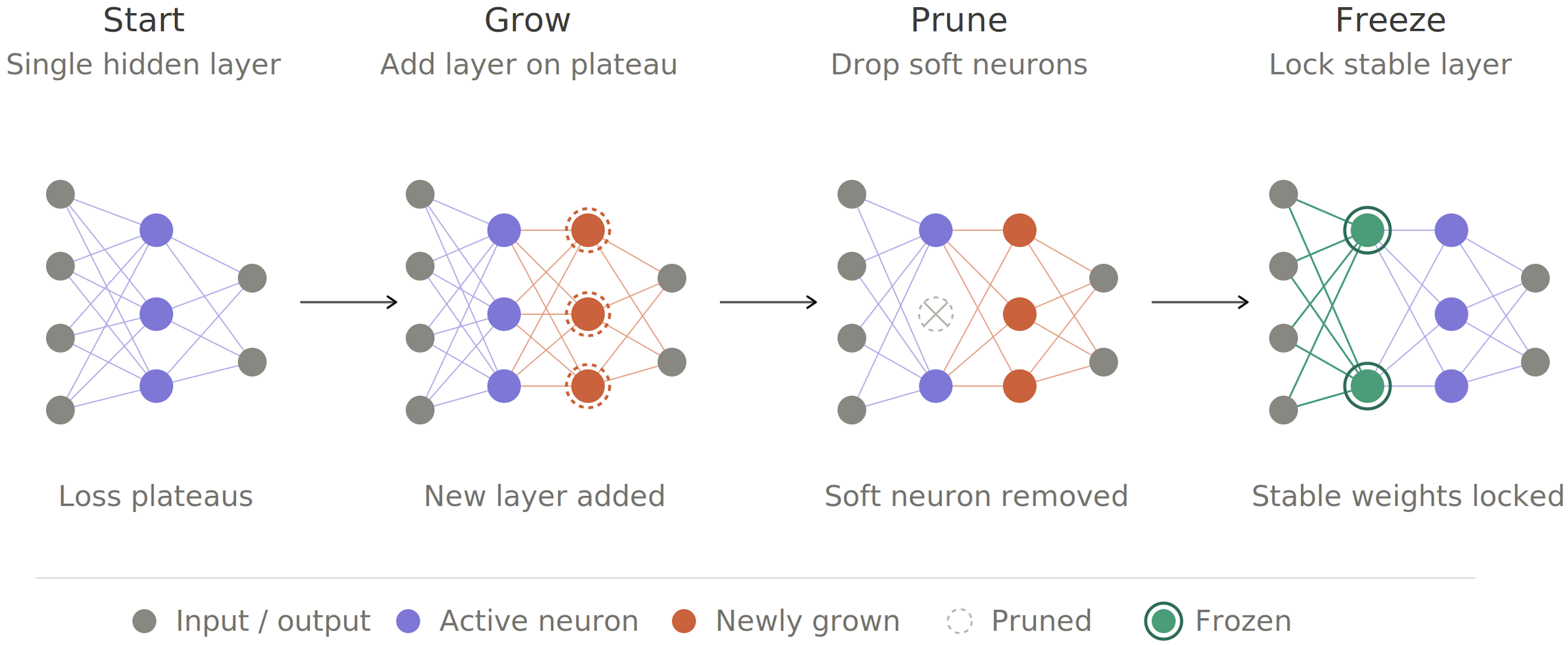}
  \caption{Conceptualization of a Grow-Prune-Freeze (GPF) neural network.}
  \label{fig:gpf_concept}
\end{figure}

Modern deep learning architectures are typically static, limiting their applicability in dynamic and non-stationary tasks. 
In addition, many of these modern models are based on transformers trained on petabytes of internet data.
The luxury of big data does not exist among the unstandardized and data-starved world of olfaction which discourages work in representative world models. 
Furthermore, molecular dynamics cause an environment to constantly evolve, and some environments are more dynamic than others which correspondingly warrant more compute in order to correlate molecular patterns within the air. 
Tasks like olfactory navigation significantly benefit from runtime learning.
As a result, we became drawn toward machine learning methods that could continuously update their own priors based on batches of observed data, and adjust their own architectures according to the observed data complexity.
These desires led us to develop \textit{Grow-Prune-Freeze (GPF) networks}, grounded in known work from random matrix theory and scaled from first-principles mathematics.

Our motivation is to develop a useful method for robotic runtime learning in partially observable world models, with a specific emphasis on olfactory navigation: a robot must locate a chemical source by following a turbulent odor plume in sparse-signal conditions using an edge-deployable policy-gradient model. 
Growing layer-by-layer enables a network to adapt capacity to plume complexity, pruning keeps the model compact for onboard inference, and freezing preserves temporal dynamics learned from early plume encounters to hedge against catastrophic forgetting.
We selected foundational temporal difference learning and pruning methods to emphasize that the performance observed in our experiments is credited to \textit{GPFs} rather than an exotic learning architecture.
In further support of \textit{GPFs} as a generalized method, we elaborate on their intuitive extension to other learning tasks in the Supplementary Material.


\section{Grow-Prune-Freeze Networks}

Continual learning requires adapting indefinitely to non-stationary streams while maintaining \emph{stability} (retention of prior knowledge) and \emph{plasticity} (ability to learn new tasks). 
Standard deep learning suffers catastrophic plasticity loss in continual settings as shown by \cite{Dohare2024_plasticity_loss}, and while adaptive pruning methods such as those in \cite{lasby2024dst,frankle2019lotterytickethypothesisfinding} show promise, they lack theoretical grounding for long-term spectral stability.

\textit{Grow-Prune-Freeze} neural networks offer a framework that automatically (1) grows hidden layers when validation loss plateaus; (2) prunes ``soft'' neurons according to temporal belief values; and (3) freezes stable layers to prevent forgetting while preserving plasticity.

We root \textit{GPFs} in random matrix theory (RMT) and analogize the mathematics to theories of deep learning architectures.
Specifically, we show that the work of \cite{Pennington2017} can be extended from single-layer neural networks to $n$-layer continuous-learning deep neural networks that incrementally add layers as they learn while preserving eigenvalue spectrality.
Through this, we also explore how a \textit{GPF's} low-rank perturbations maintain Marchenko-Pastur spectral densities with the appropriate shape, ensuring dynamical isometry and gradient propagation. 
We then show that principles from \cite{tao2012randommatrixtheory,Martin2021RMT} hold as a result.
The final model partially addresses the three plasticity correlates identified by \cite{Dohare2024_plasticity_loss}: (1) weight magnitude growth, (2) dead units, and (3) effective rank collapse.

%% file: src/_related_work.tex
\subsection{Relevant Background}

Random matrix theory is the bedrock from which \textit{GPFs} were inspired and built.
\cite{Pennington2017} derive spectral densities for non-linear activations in basic neural networks with a single hidden layer, and \cite{pmlr-v151-cohen-karlik22a} link low-rank bias to generalization. 
\cite{Martin2021RMT} cover RMT as it pertains to deep learning by showing how model width and \textit{iid} weight assumptions can be checked and controlled.
We extend these works to non-equilibrium ensembles induced through the the continual adding of layers in multilayer perceptrons.

\textit{Dynamic Sparse Training (DST)} by \cite{lasby2024dst} alternates pruning and regrowth of neural networks, while MorphNet from \cite{gordon2018morphnet} optimizes sparsity via regularization. 
Growing methods like \textit{DEN} from \cite{yoon2018lifelong} add neurons according to loss plateaus. 
General principles for constructing adaptive neural networks are recently enumerated by \cite{mirzadeh2022architecturematterscontinuallearning} and backed by empirical support on recent model architectures.

\textit{Elastic Weight Consolidation} from \cite{kirkpatrick2017catastrophicforgetting} highlights weight-specific penalization during learning; \textit{Progressive Nets} by \cite{Rusu2016Progressive, rusu2022progressiveneuralnetworks} expand layers laterally to support additional inputs. 
\cite{Dohare2024_plasticity_loss} demonstrate plasticity loss across optimizers and propose continual backpropagation (CBP) via selective reinitialization. 
Additional work from \cite{li2021neuralplasticitynetworks} and  \cite{lyle2023plasticity} denote other challenges in building adaptive architectures that continually learn.
\textit{GPFs} build on top of principles established in each of these works by using random matrix theory as the grounding ideology and \textit{Columnar Constructive Netwroks} established by \cite{javed2023_columnar_constructive_networks_ccn} as the inspiration for continually growing layers.

\cite{fehring2025growingexperiencegrowingneural} show grow-and-prune methods may be empirically effective but lack theoretical characterization of how structural adaptation affects network dynamics.
Magnitude-based pruning from \cite{lecun1989optimalbraindamage}, the Lottery Ticket Hypothesis by \cite{frankle2019lotterytickethypothesisfinding}, and parameter freezing from \cite{foster2017parameter} all give supporting evidence that sparse, stabilized networks preserve critical knowledge.
In \textit{GPFs}, pruning and freezing are integrated with growth to form a continuous cycle of adaptive architecture.
We select LeCun's magnitude-based pruning as the specific method for our evaluation.

%% file: src/_methods_a.tex
\section{Theoretical Analysis}

The \textit{GPF} framework assumes a very simple neural network to start: an input layer, an output layer, and a single hidden layer. 
Consider a feedforward network with $L_t=1$ layer at time $t=0$, input dimension $n_0$, hidden widths $n_1, \dots, n_{L_t-1}$, and output $n_{L_t}$.
Weights $W^\ell \in \mathbb{R}^{n_\ell \times n_{\ell-1}}$ are initialized from Kaiming distribution.
\textit{GPFs} grow depth only as their environment demands it, preserving the eigenvalue spectrality of previously learned weights.

To facilitate this, we define a threshold $\omega_l$ within the validation loss. 
As it learns, if the network experiences no change in loss above this threshold, a new hidden layer $\ell_i$ is added, where $i >0, i \in \mathbb{R}$.
At time $k$, there is a ``patience'' variable $\rho_k$ such that there must be $\rho_k$ epochs with validation loss all below this threshold in order to add a new layer.

The above can be denoted as follows: Add layer $\ell_i = L_t+1$ if validation loss $\mathcal{L}_\text{val}$ satisfies:
\begin{equation}
    \mathcal{L}_\text{val}^{(t-k)} - \mathcal{L}_\text{val}^{(t)} < \omega_l, \quad \forall i \in [t-k+1, t],
    \label{eq:growth}
\end{equation}
where $\omega_l > 0$ is the stagnation threshold.


\noindent Each neuron's input weight $w_{i,j}^\ell$ has belief $h_{i,j}^\ell \in [0,1)$. 
This belief value is such that if the absolute value of a neuron's weight is consistently above a positive value $\omega_w$ (or that neuron is constantly fired) for multiple successive epochs, then it is incremented, and thus becomes ``hardened''.
New layers initialize $n_{L_t+1} = n_{L_t}$ neurons with $h_{i,j}^{(L_t+1)} = 0$.
We perform network updates according to the following:
\begin{equation}
    h_{i,j}^\ell(t) = h_{i,j}^\ell(t-1) + \eta_h \cdot \mathbbm{1}\left\{ |w_{i,j}^\ell(t)| > m \right\},
    \label{eq:belief_update}
\end{equation}
where $\eta_h = 1/(k_{\textit{max}})$ normalizes to approach but never reach 1 according to the current training epoch $k_{\textit{max}}$. 
Hardened neurons are asymptotically closer to $1$ and adverse to pruning.
A belief value of $1$ is associated with a neuron considered as ``frozen'', which we discuss shortly.

Neurons with persistently high belief are hardened and resistant to pruning.
We follow the magnitude-based pruning technique established by \cite{lecun1989optimalbraindamage} with the exception that a layer's belief values protect useful low-magnitude weights that encode sparse but relevant patterns as shown by \cite{frankle2019lotterytickethypothesisfinding}.

Parameter $\omega_p$ is a layer-wise pruning threshold.
After at least $\rho_b$ epochs have occurred since the addition of layer $\ell_i$, all layers $\ell_0 - \ell_{i-1}$ prior to the one just added will be pruned such that all neurons within said layers containing belief values below some threshold of $|\omega_p|$ will be removed from the network, or their weights set to zero. 

Succinctly, we prune neuron $j$ in layer $\ell_i < \ell_k$ if:
\begin{equation}
    h_{i,j}^\ell < \omega_p \quad \text{and} \quad |w_{i,j}^\ell| < d_i.
    \label{eq:pruning}
\end{equation}

Newly added layers start with neurons whose belief values each have $h=0$ and are exempt from pruning until $\rho_b$ patience epochs have elapsed; the grow-prune cycle continues indefinitely.

Another set of tunable parameters are $\rho_f$ and $\omega_f$, where $\rho_f < k_{\textbf{max}}$ and $\omega_f \in [0,1]$.
We freeze layer $\ell_i$ if, for a patience of $\rho_f$ epochs, $100.0 * \omega_f\%$ weights satisfy:
\begin{equation}
    |w_{i,j}^\ell(k_{\textbf{max}}-q) - w_{i,j}^\ell(k_{\textbf{max}})| < \omega_f, \quad \forall q \in [k_{\textbf{max}}-\rho_f+1,k_{\textbf{max}}].
    \label{eq:freezing}
\end{equation}

Freezing follows methods given by \cite{Dohare2024_plasticity_loss} to prevent forgetting while maintaining plasticity in newer layers. Optionally, batch size may decay as $b_t = b_0\cdot\exp(-\eta_{bs}\mathcal{L}_\text{val}^{(t)})$ to account for evolving network size and convergence rate as shown by \cite{adachi2024_adaptive_batch_size}.
As we soon show, freezing stable layers also ensures that the spectral distribution of the Gram matrix is preserved.

\subsection{Training GPFs}

As the network begins training, it does not know how many layers it will eventually add.
One may cap the layer count, but we argue that a properly developed \textit{GPF} will cease growing layers at the appropriate time if designed correctly.
A continually learning model may, in some instances, maintain the same neuron count at time $t_0$ as at time $t_n$, but with more layers and increased sparsity.
We desire sparse models that grow in depth but necessarily and sufficiently parameterize themselves to the data they perceive.
In this manner, the concepts of training and testing being two separate processes are foregone.
For \textit{GPFs}, feedback is directly incorporated according to its experience or training data and there is no separate test phase.
We design \textit{GPFs} to be employed in scenarios where there is too little data for conventional training mechanisms and/or where the data may contain temporal patterns that change over time according to new dynamics introduced into the environment, as is the case with plume tracking in olfactory robotics.

\subsection{Analysis via Random Matrix Theory}
\label{sec:multilayer_rmt}

\cite{Pennington2017} showed that the distribution of eigenvalues from the data covariance matrix is idempotent after passing through a single nonlinear layer of a neural network. 
We aim to show that this principle extrapolates to any number of layers $L \in [1, \infty]$ as long as (1) stable depth-wise growth occurs within the network according to \cite{javed2023_columnar_constructive_networks_ccn}, (2) pruning is done according to the Hessian approximation in \cite{lecun1989optimalbraindamage}, and (3) layers are frozen at the appropriate time to preserve pruned knowledge.
We show how principles from \cite{Schoenholz2016} govern the limit for how deep a network can be given a set of hyperparameters, and how decisions on growth can occur based on their assumptions.

We follow the notation of \citet{Pennington2017}, reviewed in full in Supplementary Section~\ref{sec:rmt_background}. The core object at each layer $\ell$ is the empirical Gram matrix $M^{(\ell)} = \frac{1}{m}[Y^{(\ell)}]^\top Y^{(\ell)}$ where $Y^{(\ell)} = f(W^{(\ell)} Y^{(\ell-1)})$, and its Stieltjes transform $s^{(\ell)}(z) = \frac{1}{m}\operatorname{Tr}((M^{(\ell)}-zI)^{-1})$.

\textbf{We extend the above to the multi-layer case.} For a network with $L$ layers and weight matrices $W^{(1)},\dots,W^{(L)}$, define the activation recursion $Z^{(\ell)} = W^{(\ell)} Y^{(\ell-1)}$, $Y^{(\ell)} = f(Z^{(\ell)})$, $Y^{(0)} = X$, under the assumptions established by \cite{Pennington2017} (highlighted in Supplementary Section~\ref{sec:rmt_background}). The layer-$\ell$ Gram matrix and Stieltjes transform are:
\begin{equation}
    M^{(\ell)} \;=\; \frac{1}{m}\,[Y^{(\ell)}]^\top Y^{(\ell)},
    \label{eq:gpf_multilayer_gram_matrix}
\end{equation}
\begin{equation}
    s^{(\ell)}(z) \;=\; \frac{1}{m}\operatorname{Tr}\!\bigl( (M^{(\ell)}-zI)^{-1}\bigr).
    \label{eq:gpf_multilayer_gram_matrix_stieltjes}
\end{equation}

The nonlinearity of activation functions precludes many principles from random matrix theory applying to neural networks.
While many of the tools for assessing moments and spectral densities from matrix theory still apply, we show how these tools can be further employed to analyze space folds with ReLU activations (and friends) as demonstrated by \cite{lewandowski2025onspacefoldsrelu}.

Let $\Psi_\ell$ be an activation function that acts as a \emph{nonlinear spectral transformer}, encoding how a random layer reshapes the spectral statistics of the preceding layer.  
Explicit expressions for $\Psi_\ell$ depend on integrals involving the activation $\psi \in \Psi$ and may reduce to polynomial or rational forms for common nonlinearities.
Under the moments-method derivation (counting non-linear random walks at each layer) one obtains, for each \(\ell\), a deterministic operator \(\Psi_\ell\) 
that enables Equation \ref{eq:gpf_multilayer_gram_matrix_stieltjes} to be written as
\[
\qquad
s^{(\ell)}(z) \;=\; \Psi_\ell\!\bigl(s^{(\ell-1)}(z),\, z\bigr)
\qquad
\]
for every \(z\in\mathbb{C}\setminus\mathbb{R}\), where \(\Psi_\ell\)  depends on the activation \(\psi\), layer variance \(\sigma_\ell^2\).
Thus, when depth is increased by adding layers, the transform is updated by \emph{composition} of these operators. 
We can define the depth-\(L\) composition operator
\[
\Psi^{(L:1)} \;=\; \Psi_L\circ\Psi_{L-1}\circ\cdots\circ\Psi_1,
\]
so that the Stieltjes transform at depth \(L\) is succinctly written as

\begin{equation}
    s^{(L)}(z) \;=\; \Psi^{(L:1)}\!\bigl( s^{(0)}(z),\, z\bigr).
    \qquad
    \label{eq:stieltjes_depth_L}
\end{equation}



We formalize the conditions for convergence of GPFs at any layer depth limit via Theorem \ref{thm:depth-limit}.

\begin{theorem}[Depth-Limit under Composition]
\label{thm:depth-limit}
Suppose the layers are homogeneous such that 
\(\psi_\ell = \Psi \forall \psi \in \Psi\).  
Assume there exists an open set \(\mathcal{U}\subset\mathbb{C}\setminus\mathbb{R}\)
containing the spectral support of \(M^{(\ell)}\) for all \(\ell\), and a Banach 
space \((\mathcal{S},\|\cdot\|)\) of analytic transforms on \(\mathcal{U}\) such 
that:

\begin{enumerate}
\item Condition 1: \(\Psi:\mathcal{S}\times\mathcal{U}\to\mathcal{S}\) is well-defined and 
analytic in its first argument for each fixed \(z\in\mathcal{U}\).
\item Condition 2: For each fixed \(z\in\mathcal{U}\), the Fréchet derivative 
\(D_s\Psi(s,z)\) exists and there is a constant \(\rho(z)<1\) with
\[
\sup_{s\in\mathcal{S}}\|D_s\Psi(s,z)\| \le \rho(z) < 1.
\]
\end{enumerate}

Then for every initial Stieltjes transform \(s^{(0)}\in\mathcal{S}\), the iterates
\(\{s^{(L)}\}_{L\ge 0}\) defined by \(s^{(L)}=\Psi^{\circ L}(s^{(0)},\cdot)\)
converge in \(\mathcal{S}\) to the unique fixed point \(s^\star(\cdot)\) that 
satisfies \(s^\star(z)=\Psi(s^\star(z),z)\) for all \(z\in\mathcal{U}\).
Consequently, the empirical spectral distribution of \(M^{(L)}\) converges 
(as \(m,n_0,\dots,n_L\to\infty\), jointly with \(L\to\infty\)) to the 
deterministic law given by the Stieltjes transform \(s^\star\).
\end{theorem}




Freezing layers locks spectral distributions into their ideal shape to further enable convergence.
Our analysis assumes layer homogeneity, with new layers always initialized at identical width $d$.
Formal convergence results along with further remarks on Theorem \ref{thm:depth-limit} and ReLU spectral analysis appear in Supplementary Sections~\ref{sec:homogeneous_layers} and \ref{sec:ch14_relu}.







%% file: src/_experiments_a.tex
\section{Experiments}
\label{sec:experiments}

\begin{figure}[h]
  \centering
  \includegraphics[width=\linewidth]{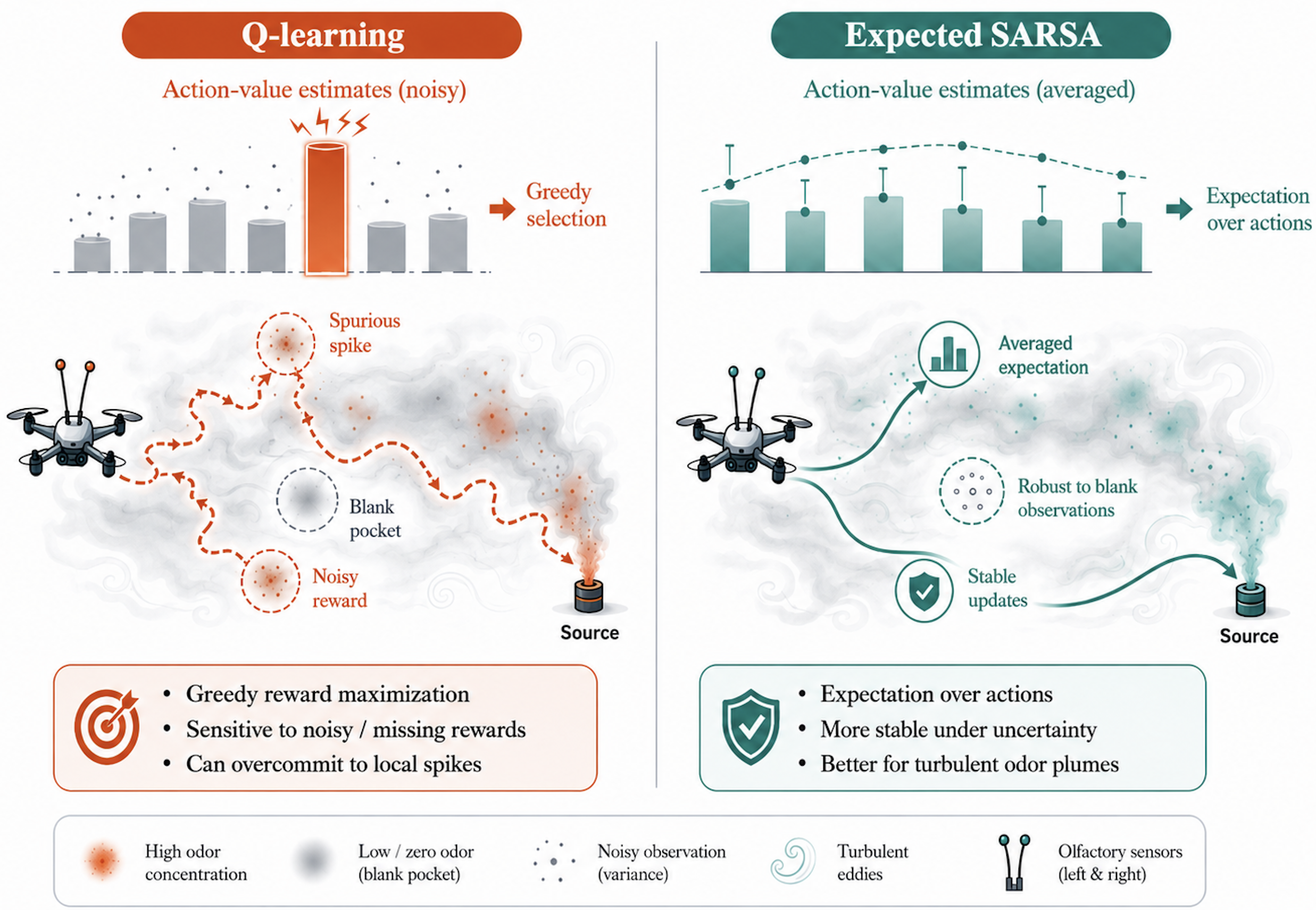}
  \caption{Q-learning vs Expected SARSA in Olfactory Navigation}
  \label{fig:q-learn_vs_expected_sarsa}
\end{figure}


We evaluate the \textit{GPFs} on olfactory navigation - the problem of a robot localizing an odor source by following a turbulent plume back to its origin.
This task is representative of the class of sparse-reward, partially observable navigation problems that motivate runtime learning algorithms in big worlds: the signal is intermittent, the environment is non-stationary, and the appropriate behavioral strategy depends on context that must be inferred from current state and recent sensory history.

\subsection{Plume Simulator}

We simulate the plume using a Farrell--Murlis filament model~\cite{Farrell2002FilamentBasedADPlume} in a $20\,\text{m}\times 20\,\text{m}$ two-dimensional domain at $10\,\text{Hz}$. 
The source emits filaments at a Poisson-sampled $\lambda = 5\,\text{filaments/s}$ from a randomly chosen upwind position. 
Filaments advect with a coupled Ornstein--Uhlenbeck wind field with velocities of $1.0 \text{m/s}$, gusting durations of $2.0\text{m/s}$, and spread via Gaussian diffusion rate of $0.05\,\text{m}^2/\text{s}$
Total concentration is clamped to $1.0$ and corrupted by additive noise $\sigma_n = 10^{-3}$. 
Full details on the simulator construction are enumerated in Supplementary Section~\ref{sec:simulator_details}.

\subsection{Agent and Sensor Model}

The agent carries a bilateral antenna pair ($10\,\text{cm}$ inter-antenna separation) and selects from six discrete actions: \textit{surge forward}, \textit{turn} $\pm 15^\circ$, \textit{turn around} $180^\circ$, and \textit{cast} $\pm 30^\circ$, each followed by a $50\,\text{cm}$ forward step.
Olfactory inertial odometry (OIO) established by \cite{france_2025_oio_method} is used to fuse inertial and olfactory states.
The antennae are modeled as fast-sampling Sensirion MICS-6814 metal oxide sensors building on those models established by recent olfactory navigation work from \cite{dennler_high-speed_2024} and \cite{france2026_ghosts_mdpi}.
Episodes terminate when the agent moves within 50 cm of the source or after $1200$ steps ($100\,\text{s}$). 
Initial placement of the odor source is $3$--$10\,\text{m}$ downwind with $\pm 30^\circ$ heading noise.
Sensor observations are discretized into a $22$-dimensional state: seven concentration bins (blank $+$ six log-spaced levels above $\sigma_n$) per antenna and eight wind-direction octants relative to heading, giving $7{\times}7{\times}8 = 392$ unique states.
Further details on state- and action-space tokenization can be observed in Section \ref{sec:simulator_details} of the Supplementary Material.

\subsection{Expected SARSA Policy}
We build off the work established by \cite{France2023} and \cite{france2026_ghosts_mdpi} showing the advantage Expected SARSA can provide against greedy Q-learning in partially observable environments with sparse rewards.
We find that a policy that seeks the maximum reward may induce lossy actions when rewards are sparse and not fully observable; a policy that seeks the \textit{average} reward for a given action set supports robustness against eddy currents and blanks in a plume.
Figure \ref{fig:q-learn_vs_expected_sarsa} illustrates this comparison between Q-learning and Expected SARSA agent policies.

Furthering this hypothesis, we select the latter as the navigation policy for both a canonical agent and \textit{GPF} agent.
The canonical agent contains a single hidden layer to evaluate as a control against the \textit{GPF} agent.
Both agents are identical in architecture otherwise.
The Q-function is represented by a simple dense MLP with ReLU activations.  
The initial architecture is $22 \to 64 \to 6$ (one hidden layer of width $64$) trained with the Expected SARSA update rule under an $\varepsilon$-greedy policy, with $\varepsilon$ decayed from $1.0$ to $0.05$ multiplicatively by $0.9995$ per episode.  
We select the Adam optimizer with learning rate $\alpha = 1e^{-3}$, and a discount factor $\gamma = 0.99$.
We keep the neural network basic for two reasons: (1) our intent is to evaluate \textit{GPF} methodology, not an exotic network architecture, and (2) a simple network ensures maximum exportability to and reproducibility in various model interpreters required by edge robotics.
The belief-value pruning threshold is set as $\omega_p = 10^{-6}$.



The per-step reward $r_t$ for both agents combines a constant time penalty of $r_{\text{time}} = -0.01$ with event signals $r_{\text{event}}$ of $+100$, $+1.0$, and $-0.05$ for success, whiff, and prolonged blanks $> 2\,\text{s}$, respectively.
A potential-based distance-shaping term $r_{\text{shape}}$ \citep{Ng1999rewardshaping} is delegated to provide dense gradient signal without altering the optimal policy.
In culmination of the above, the final reward function is represented as  $r_t = r_{\text{time}} + r_{\text{event}} + r_{\text{shape}}$.
All experiments use seed $42$ for the environment and seed $43$ for the agent.  
Training runs for $4{,}500$ episodes with evaluation every $500$ episodes over $200$ held-out episodes.  
The highest-performing checkpoint is retained. 
Full hyperparameters for all experiments are available in Table~\ref{tab:gpf_hyperparams} of the Supplementary Material.

%% file: src/_results.tex

\section{Results}

\begin{figure}[h]
  \centering
  \includegraphics[width=\linewidth]{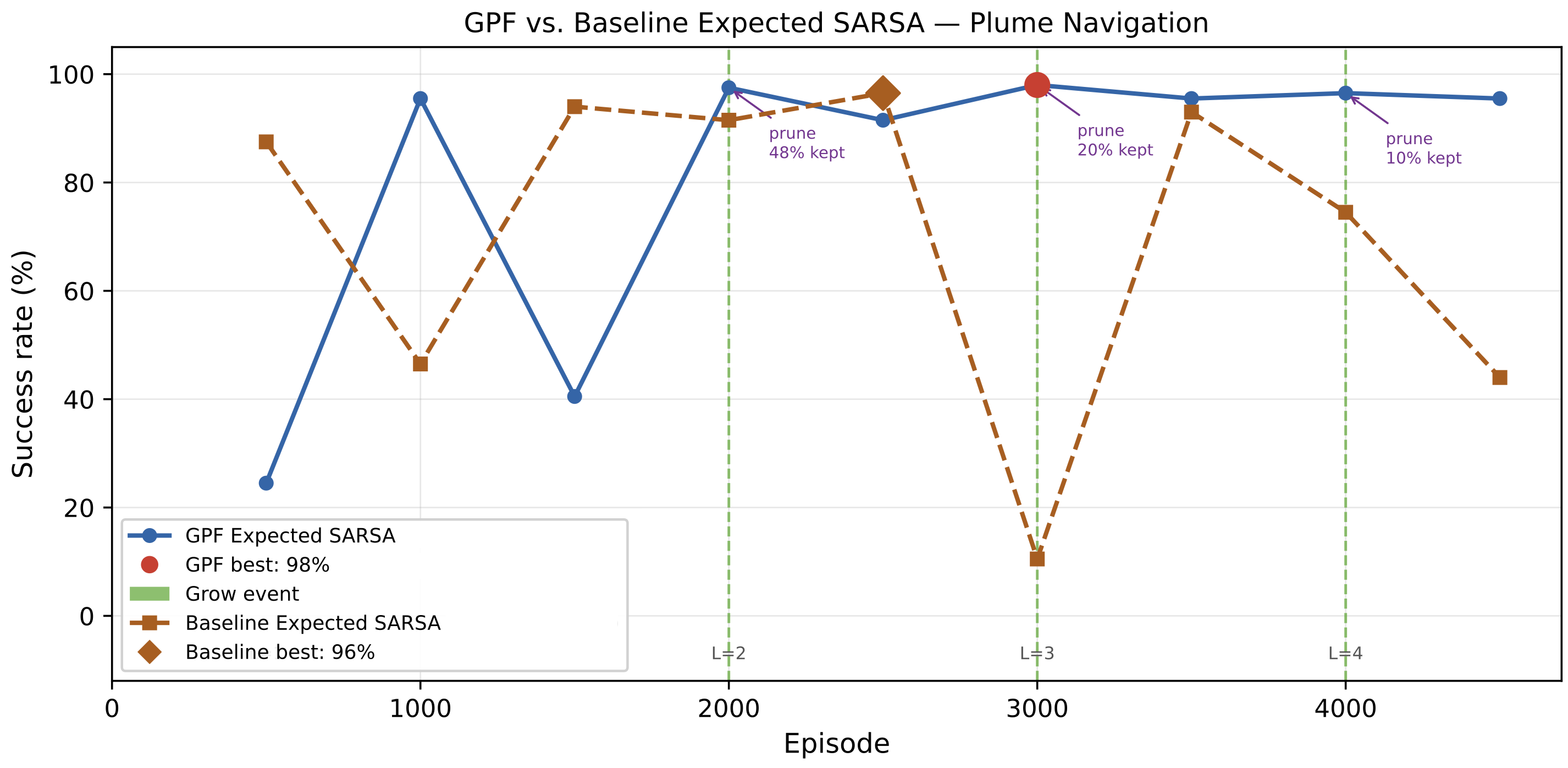}
  \caption{\textit{GPF} Expected SARSA training curve on the plume navigation task.
           The orange dotted lines denote the success rate of the baseline agent; the solid blue line denotes that of the GPF agent. Dashed green lines mark grow events; purple annotations show the fraction of weights retained by each prune pass. The red dot marks the best checkpoint. The bottom access indicates layer count.}
  \label{fig:gpf_training_curve}
\end{figure}

Figure~\ref{fig:gpf_training_curve} shows the evaluation success rate throughout training. 
The agent requires roughly $1000$ episodes to acquire consistent navigating behavior. 
Three grow events occurred at episodes $2{,}000$, $3{,}000$, and $4{,}000$. 
After the first grow-prune pass (retained $47.9\%$ of weights), the two-layer network temporarily dropped to $91.5\%$ before recovering to $\mathbf{98.0\%}$ at episode $3{,}000$ - the best checkpoint. 
Subsequent grows were more aggressive ($19.6\%$ and $9.9\%$ weight retention); the four-layer configuration peaked at $95.5\%$ before degrading to $68.0\%$ by episode $5{,}000$ as $\varepsilon$-decay compounded ultra-sparse representation. 
The full episode-by-episode trajectory is in Table~\ref{tab:gpf_training_curve} of the Supplementary Material.
Extending the episode budget from $50\,\text{s}$ to $100\,\text{s}$ raised the peak from $94.5\%$ to $98.0\%$ by eliminating timeout failures on hard initializations where the agent is far downwind or in a weak-signal region, allowing full surge--cast--surge cycles to complete as necessary.


Evaluating this checkpoint over $100$ independent episodes with distinct random seeds, the agent achieved $94\%$ success in acquiring the source of the plume. 
All six failures were timeouts in which the agent correctly tracked the plume but exhausted the episode budget - consistent with timing rather than navigational strategy as the remaining failure mode.
We do not increase the episode budget because the molecular dynamics of the odors of interest would equalize within the plume beyond the allotted time, leaving no signal for the agent to trace.
We leave varying episode budgets according to chemical diversity and plume density as an intuitive avenue for future work.


The GPF agent learned a high-performance navigation policy through structured growth of a sparse four-layer Expected SARSA policy that achieved $98\%$ on the training distribution and $94\%$ on held-out random initializations. 
These results support that the GPF grow-and-prune mechanism successfully compresses a capable policy into a sparse architecture, and that the 100-second episode budget is sufficient for the agent to develop and execute the multi-phase olfactory search strategies - surge, cast, and loiter - that characterize successful biological plume navigation.

To provide support that \textit{GPFs} generalize beyond olfactory navigation, we perform a battery of experiments over standard image classification, Atari reinforcement learning, and autoregressive language modeling problems. These results can be found in Section \ref{sec:gpf_general_experiments} of the Supplementary Material.

%% file: src/_future_work.tex
\section{Limitations \& Ongoing Work}
\vspace{-3mm}
\begin{figure}[!b]
  \centering
  \includegraphics[width=\linewidth]{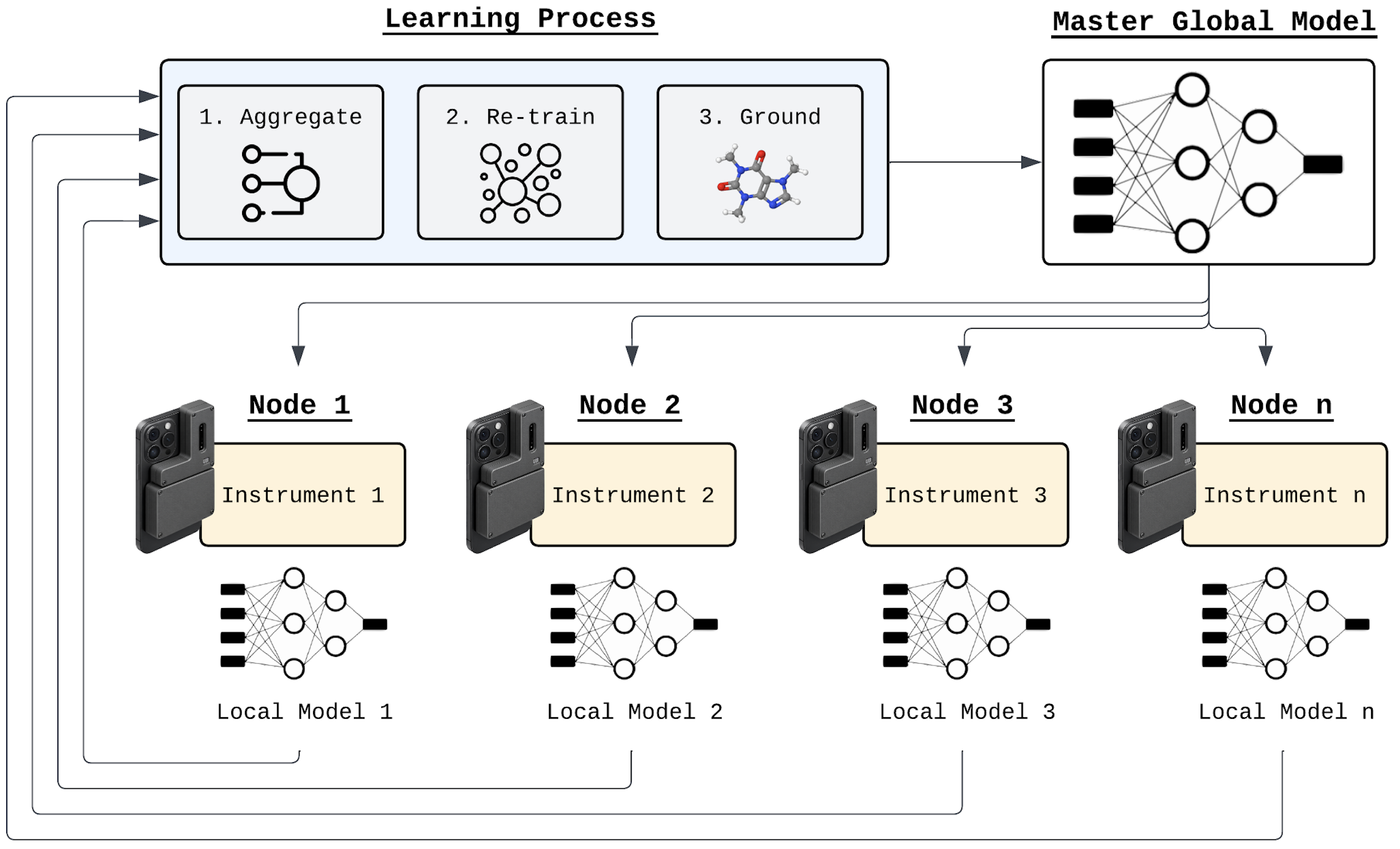}
  \caption{The federated learning mechanism for the \textit{Sigma} olfaction-vision-language model.}
  \label{fig:sigma_fl}
\end{figure}

We built \textit{GPFs} out of necessity for effective on-the-fly learning specifically for olfactory navigation tasks.
While we strived to design a practical and grounded method for temporal difference learning at runtime that also generalizes well to other tasks, we acknowledge that there are still many more empirical evaluations to be done.
One intuitive avenue of future work is to extend the simulator to handle more complex gas dynamics, advanced molecular interactions, additional sensor models, and longer wind events.
An exhaustive simulator would require a full computational fluid dynamics (CFD) engine.
However, we feel that effective olfactory navigation algorithms will not require a granular onboard CFD engine, but rather benefit from a thorough understanding of first-order approximations of molecular dynamics according to the target compound being tracked.

Recent work from \cite{jolicoeurmartineau2025trm} and \cite{baek_bengio2026generativerecursivereasoning} have highlighted the advantage of refining a latent vector by recursively reasoning through a single hidden layer.
In a similar manner, the architectural simplicity of \textit{GPFs} allows for recursive calls of the same hidden layer - since all layers are identical, a machine learning model need only define one hidden layer, and can just load the $i^{th}$ indexed weights for the $i^{th}$ recursion call.
Computational complexity remains the same, but size complexity may be noticeably smaller in very deep architectures that conventionally export layers sequentially.
This becomes most prudent on edge devices and robots that need to make decisions with onboard compute.
Ongoing research is evaluating the potential of such a mechanism.

Federated learning poses as another worthy domain for integration of \textit{GPFs}.
Leveraging de-centralized learning techniques such as those established by \cite{bao2025_federated_learning_clip} and \cite{pmlr-v54-mcmahan17a-federatedlearning} could allow faster and assured convergence of eigenvalue spectra among model weights by enabling multiple \textit{GPF} models to learn from each other.
Such methods could also automate hyperparameter tuning for pruning and freezing neurons.
Initial investigations of federated learning among \textit{GPFs} in the context of olfaction-vision-language models embedded on mobile devices are currently underway with our work in \cite{scentience2025_ovlm} and \textit{Sigma} from \cite{sigma_france}.
A diagram for the \textit{Sigma} architecture can be found in Figure \ref{fig:sigma_fl}, where each ``Local Model'' defines a series of multimodal \textit{Grow-Prune-Freeze} neural networks within a quantized mixture-of-experts model.

%% file: src/_conclusion.tex
\section{Conclusion}
\label{sec:conclusion}

\vspace{-3mm}
\textit{GPF} networks were developed out of a need for lightweight and adaptable olfactory navigation agents that could be deployed at the edge on real robots. 
Olfaction does not have the luxuries of other modalities in plentiful and accessible world models.
As such, plume navigation tasks significantly benefit from adaptable runtime learning.
\textit{GPFs} and temporal difference learning provide an encouraging solution for this by growing depth according to observed environment complexity, prune low-saliency weights as learning occurs, and freezing stable layers - addressing catastrophic forgetting and plasticity loss while maintaining spectral invariance via Stieltjes composition.

On the olfactory navigation task, the 94\% source-finding success rate supports GPFs as a practical candidate for adaptive temporal difference learning agents operating in partially observable, non-stationary environments, while also gathering evidence as a continual learning framework for the same.
A larger experimental battery across diverse benchmarks and chemical plumes remains the primary open direction to validate \textit{GPFs} on a more comprehensive level.
We hope our work here inspires further research into continual learning and olfactory world models for robotics.

All code for the experiments associated with this paper is available at the following URL: 
\url{https://github.com/KordelFranceTech/Grow-Prune-Freeze-Neural-Networks}

%% file: src/_impact_statement.tex
\subsubsection*{Broader Impact Statement}
\label{sec:broaderImpact}

Our work here progresses advanced navigation abilities for robots.
Olfaction is still a largely unexplored modality.
It lacks a unifying data standard, ImageNet-scale training datasets, and no formal benchmarks for tasks like scent-based navigation.
Maturing olfaction to the degree of its vision and audition counterparts will enable many new capabilities for AI and robotics.
However, these will be capabilities for which humanity has little experience in addressing due to the aforementioned obstacles.
Additionally, enabling AI models to adapt their own architectures and continually learn as we demonstrate here with \textit{GPFs} could enable unintended behaviors.
A \textit{GPF} model may be deployed against certain criteria and alter itself to learn action sequences adversarial to its original policy.

Consequently, we feel our work here adds to the growing call-to-action to establish meaningful multimodal benchmarks for continual learning to ensure adaptive models remain grounded as they explore the world. 
We strongly encourage the olfactory modality to be included in this benchmark.


%% file: src/supplementary/_framework_table.tex
\section{GPF Hyperparameter Reference}
\label{sec:framework_table}

\begin{table}[h]
    \caption{GPF Network Hyperparameters}
    \begin{center}
    \begin{tabular}{|c|c|}
    \hline
    \textbf{Symbol}& \textbf{Description}\\
    \hline
    $k_{\textbf{max}}$& Maximum training epochs\\
    $d_i$& $i^{th}$ layer dimension\\
    $\rho_e$& Epoch patience required before GPF is enabled\\
    $\rho_k$& Epoch patience required before growing a layer\\
    $\rho_p$& Epoch patience required before pruning neurons\\
    $\rho_f$& Epoch patience required before a neuron can be frozen\\
    $\omega_v$& Loss stagnation threshold for layer growth\\
    $\omega_p$& Belief threshold for pruning\\
    $\omega_f$& Weight-change threshold for layer freezing\\
    $\alpha$& Learning rate\\
    $\gamma$& Discount factor\\
    $\epsilon$& Exploration greediness\\
    $\eta_g$& Greediness decay\\
    $\eta_{bs}$& Batch size decay\\
    $\eta_h$& Belief value increment\\
    \hline
    \end{tabular}
    \label{tab:paramsTable}
    \end{center}
\end{table}

%% file: src/supplementary/_pseudocode.tex
\section{GPF Pseudocode}
\label{sec:pseudocode}

In consolidation of the formal logic introduced above, one can succinctly define how the GPF network evolves according to the following rules:

\begin{enumerate}
    \item \textbf{Layer Growth:} Add a new hidden layer if validation loss stagnates over $\rho_v$ epochs:
    \[
    \Delta L_\text{val}^{(t-k:t)} < \omega_v
    \]
    \item \textbf{Neuron Belief Hardening:} Each neuron $i$ has belief $h_i \in [0,1)$, incremented if
    \[
    |w_i| > \omega_v \text{ over } k_{\textbf{max}} \text{ epochs}.
    \]
    \item \textbf{Pruning and Layer Freezing:} Neurons with $|w_i|<\omega_v$ and $h_i<\omega_p$ are pruned. Layers with $>100.0 \times \omega_f\%$ stable weights over $\rho_f$ epochs are frozen.
\end{enumerate}

\noindent Below is the pseudocode quantifying the above rules into a manner that can be executed for the experiments in the following sections.


\begin{algorithm}[t]
\caption{Grow-Pruen-Freeze Network}
\begin{algorithmic}[1]
\State $\Omega \gets$ initialize network with single hidden layer
\State randomize weights of $\Omega$ via Glorot init
\State initialize all $h$-values of $\Omega$ to zero
\For{each epoch $t$ or until termination criteria}
    \State train $\Omega$ on batch $b$
    \State update $\Omega$ weights and $h$-values
    \If{$t > \text{patience}$}
        \If{validation loss stagnates for $k$ epochs}
            \State add new hidden layer
        \EndIf
    \EndIf
    \State prune neurons with low $h$ and small weights
    \State freeze stable layers satisfying criteria $v$
    \State update $h$-values
\EndFor
\end{algorithmic}
\end{algorithm}
\newpage

%% file: src/supplementary/_rmt_background.tex
\section{Single-Layer RMT Background}
\label{sec:rmt_background}

We review the single-layer random matrix theory setup of \citet{Pennington2017} upon which our multi-layer extension is built.

Let \( X \in \mathbb{R}^{n_0 \times m} \) be a data matrix with \(m\) data samples, and let
\( W \in \mathbb{R}^{n_1 \times n_0} \) be a random weight matrix with i.i.d.\ entries:
\[
W_{ij} \sim \mathcal{N}\!\left(0,\, \frac{\sigma_w^2}{n_0}\right).
\]
Define the pre-activation matrix $Z = WX$ and apply a pointwise nonlinearity $f$ to obtain the post-activation matrix $Y = f(Z)$.

The empirical Gram matrix is:
\[
M = \frac{1}{m}\, Y^\top Y \in \mathbb{R}^{m \times m}.
\]
The main object of interest is the \textit{limiting spectral distribution (LSD)} of $M$ as $n_0, n_1, m \to \infty$ at comparable rates.
By the \textit{moments method} from random matrix theory, one characterizes the asymptotic eigenvalue density via the resolvent
\[
R(z) = (M - zI)^{-1},
\]
and the Stieltjes transform $s(z) = \frac{1}{m}\operatorname{Tr}(R(z))$.
In the single-layer case, this Stieltjes transform satisfies a self-consistent equation determined by the nonlinearity $f$ and the ratio $n_1/m$.
\citet{Pennington2017} showed that for common activations (including ReLU), the resulting LSD follows a Marchenko-Pastur-type law, and that weight matrices satisfying $\sigma_w^2 = 1$ yield a shape parameter $\psi \approx 1$, ensuring dynamical isometry and stable gradient propagation.
Our multi-layer extension in Section \ref{sec:multilayer_rmt} generalizes this single-layer picture to networks that grow adaptively during training.

%% file: src/supplementary/_homogeneous_layers.tex
\section{Notes on Convergence of Homogeneous GPFs}
\label{sec:homogeneous_layers}
Proceeding under this assumption, suppose each layer is identically distributed with the same aspect ratios and variances, 
so \(\psi_\ell\equiv\Psi\) for all \(\ell\in Z\).  
Then by taking Equation \ref{eq:stieltjes_depth_L} above, we have:
\[
s^{(L)}(z) \;=\; \underbrace{\psi\circ\psi\circ\cdots\circ\psi}_{L\ \text{times}}
\!\bigl(s^{(0)}(z),z\bigr) \;=\; \Psi^{\circ L}\bigl(s^{(0)}(z),z\bigr).
\]
Two immediate consequences follow:

\begin{enumerate}
\item \textbf{Monotone / contractive iteration.} If \(\psi(\cdot,z)\) is a contraction on a suitable subset \(\mathcal{S}\) of analytic transforms (for example, relative to the supremum norm on compact \(z\)-domains), then the sequence \(s^{(L)}(z)\) converges geometrically in \(L\) towards the unique fixed point \(s^\star(z)\) solving \(s^\star=\Psi(s^\star,z)\).
\item \textbf{Operator accumulation.}
Even when \(\Psi\) is not strictly a contraction globally, one may obtain convergence for initial conditions \(s^{(0)}(z)\) in a basin of attraction; numerically this corresponds to the spectral measure ``forgetting'' low-rank or sample-specific components as depth increases.
We note, however, that this ``forgetting'' is partially countered by the layer-specific freezing process defined above.
\end{enumerate}

\subsection{Remarks on Fixed-Point Depth-Limit Theorem}

We provide a few remarks on Theorem \ref{thm:depth-limit} that address its applicability to heterogeneous models (deep neural networks whose layers do not all have the same initial input dimension).





\noindent\textbf{Remarks.}
\begin{itemize}
\item Condition 2 is a sufficient contraction-style condition; weaker conditions (e.g., spectral radius \(<1\) only on a subset or average contractivity) can also be used in applications and are often verifiable by bounding the nonlinear moments that define \(D_s\Psi\).
\item For inhomogeneous layers one obtains
\(
s^{(L)}=\Psi_L\circ\cdots\circ\Psi_1(s^{(0)})
\)
and conclusions about convergence require joint control of the products of derivatives \(\prod_{\ell=1}^L D_s\Psi_\ell\).  A sufficient criterion is that there exist \(\rho_\ell\) with \(\sup_\ell \rho_\ell<1\) and 
\(\|D_s\Psi_\ell\|\le \rho_\ell\), which yields geometric decay of perturbations.
However, we leave the formal analysis of nonhomogeneous layers for future work.
\item From a combinatorial (moments) perspective, adding depth extends the class of contributing walks by allowing arbitrarily many layer-to-layer transitions; the operator \(\Psi\) summarizes the net effect of summing those additional contributions at each depth increment.
\end{itemize}

%% file: src/_methods_b.tex
\section{ReLU Activations and Spectral Density}
\label{sec:ch14_relu}

Many modern deep learning architectures make use of the rectified linear unit (ReLU) activation function, or some derivative thereof.
The performance garnered through the implementation of ReLU is well known and we intentionally constrain the use of activation functions within hidden layers strictly to ReLU and ReLU variants (e.g. GeLU, SiLU).
The reasoning for this is two-fold: constraining the hidden layers to one activation function allows me implement a form of control while ablating other architectural parameters; secondly, the transforms over the data that occur due to the non-linearities imposed by ReLU offer several points of interesting analysis with respect to \textit{GPFs}.

When the layers are ReLU or bounded activations, these integrals are often explicit or numerically tractable, enabling concrete verification of the above stability conditions.
The work of Lewandowsi, et al. \cite{lewandowski2025onspacefoldsrelu} note how ReLU networks transform input space in a piecewise-linear manner, producing folded, non-convex geometries across layers. 
Each fold corresponds to a region where the activation pattern changes, and consecutive layers compound these folds, increasing expressivity while also affecting the distribution of activations and gradients.

In the context of GPFs, adding a new hidden layer introduces additional ReLU folds, effectively increasing the number of non-linear regions in the network. 
This operation changes the activation covariance matrix and the Jacobian, which are central objects in RMT. 
Consequently, the Gram matrix $(G^{(l)}*t = \frac{1}{n*{l-1}} (Y^{(l)}_t)^\top Y^{(l)}_t)$ experiences shifts in its spectrum, potentially increasing the spectral width and introducing new spikes corresponding to strongly activated neurons.

GPF neurons that become hardened due to high belief values $(h_i)$ tend to preserve specific folds over successive epochs. 
In spectral terms, this generates persistent low-rank structures in the activation covariance, aligning with the “spiked” eigenvalues studied in RMT. 
These spikes can be interpreted as geometrically meaningful folds that represent robust features in the input space.

Pruning neurons with low activity or low belief selectively removes folds that contribute little to the effective expressivity of the network. 
Spectrally, this reduces bulk eigenvalues associated with inactive directions, sharpening the separation between critical spikes and the background, thus enhancing signal-to-noise ratios in the network’s representation.

Finally, freezing layers preserves existing folds while allowing newly added layers to introduce additional complexity. 
From an RMT perspective, freezing stabilizes the spectral distribution of weights of previously learned features, preventing catastrophic drift of critical eigenvalues while maintaining plasticity in subsequent layers.


Adaptive architectures like GPFs can dynamically allocate representational capacity to active plume regions while pruning inactive neurons. 
Hardened neurons preserve critical correlations, and frozen layers prevent catastrophic forgetting. 
We hypothesize that this mechanism improves efficiency, plasticity, and robustness in sensory-limited, dynamic environments such as olfactory robotics, which we mentioned was our primary motivation for the GPF design.
Growing the network accounts for plume complexity, pruning it adaptively filters out spurious time-correlated events, and freezing layers captures time-dependent patterns at different parts of the plume.

%% file: src/supplementary/_methods_c.tex
\section{Width vs Depth}
Research from Mirzadeh, et al. \cite{mirzadeh2022architecturematterscontinuallearning} define a key set of architectural boosters for continual learning that we found worked well in practice.
Namely, limiting the use of global average pooling and replacing it with max pooling and batch normalization seemed to limit plasticity loss as they claim.
Where we digress, however, is the argument that wider and shallower networks promote more stable continual learning, as we did not see this in our experiments.
Nearly every experiment we ran showed that deeper networks (albeit sparse after pruning) outperformed the shorter, wider networks.

The work from Mirzadeh, et al. is evidently correct, and we think the difference in experimental observations can be drawn from our training data size.
In most of our experiments, we were modeling with a small amount of training data which seems to benefit from the added depth.
If our applications had more plentiful training data, we expect that our results would converge toward theirs and we would reach the same conclusions regarding the benefits of width.

We note this as a point for future research.

%% file: src/supplementary/_simulator_details.tex
\section{Plume Simulator Details}
\label{sec:simulator_details}

The plume environment uses a Farrell--Murlis filament model~\cite{Farrell2002FilamentBasedADPlume} in a $20\,\text{m}\times 20\,\text{m}$ two-dimensional domain at $10\,\text{Hz}$ ($\Delta t = 0.1\,\text{s}$).

\paragraph{Source and filament emission.}
At the start of each episode the source is placed on the upwind half of the domain at a randomly sampled position biased toward the upwind edge ($\pm 25\%$ of domain width from center, with an additional $\pm 15\%$ jitter). The source emits discrete filaments at mean rate $\lambda = 5\,\text{filaments/s}$, drawn from a Poisson process so that $n_{\text{new}} \sim \text{Poisson}(\lambda\,\Delta t)$ filaments are released each step. Each filament is initialized at the source with positional jitter $\mathcal{N}(0,\sigma_0^2)$, $\sigma_0 = 0.01\,\text{m}$.

\paragraph{Filament dynamics.}
Filaments advect with the local wind velocity $(v_x, v_y)$ and grow by molecular diffusion. The Gaussian radius of filament $i$ at age $\tau_i$ is
\begin{equation}
  \sigma_i(\tau_i) = \sqrt{\sigma_0^2 + 2\,D\,\tau_i},
  \label{eq:filament_diffusion_supp}
\end{equation}
with diffusion rate $D = 0.05\,\text{m}^2/\text{s}$. Filaments are removed after a decay time of $30\,\text{s}$ or upon leaving the domain boundary.

\paragraph{Concentration measurement.}
The concentration contributed by filament $i$ at sensor position $\mathbf{p}$ is
\begin{equation}
  c_i(\mathbf{p}) = \frac{m}{2\pi\sigma_i^2}\exp\!\left(-\frac{\|\mathbf{p} - \mathbf{x}_i\|^2}{2\sigma_i^2}\right),
  \label{eq:filament_conc_supp}
\end{equation}
where $m = 0.1$ is the filament mass and $\mathbf{x}_i$ is the filament center. Filaments beyond $3\sigma_i$ are excluded for efficiency. Total concentration is clamped to $1.0$ and corrupted by additive noise $\mathcal{N}(0,\sigma_n^2)$ with $\sigma_n = 10^{-3}$.

\paragraph{Wind model.}
Wind speed $v$ and direction $\theta$ follow coupled discrete-time Ornstein--Uhlenbeck processes with mean-reversion coefficient $\alpha = \Delta t/\tau_c$:
\begin{align}
  v_{t+1}      &= v_t - \alpha(v_t - \bar{v}) + \sigma_v\sqrt{2\alpha}\,\epsilon_v, \\
  \theta_{t+1} &= \theta_t - \alpha(\theta_t - \bar{\theta}) + \sigma_\theta\sqrt{2\alpha}\,\epsilon_\theta,
\end{align}
where $\epsilon_v, \epsilon_\theta \sim \mathcal{N}(0,1)$ are independent, $\bar{v} = 1.0\,\text{m/s}$, $\bar{\theta} = 0\,\text{rad}$, $\sigma_v = 0.2\,\text{m/s}$, $\sigma_\theta = 15^\circ$, and correlation time $\tau_c = 2\,\text{s}$. Wind speed is clamped to a minimum of $0.1\,\text{m/s}$.

\subsection{Observation Tokenizer}
\label{sec:tokenizer_details}

Each $10\,\text{Hz}$ sensor observation is mapped to a compact discrete
representation by a fixed tokenizer before it is consumed by either the RL
agent or the generative world model.  The tokenizer defines a
domain-specific vocabulary over three independent streams - left antenna
concentration, right antenna concentration, and relative wind direction - and
encodes each stream as a single integer index.

\paragraph{Concentration quantization.}

Raw concentration readings are quantized into seven discrete levels.
Level 0 (\textsc{blank}) captures all readings at or below the sensor noise
floor; levels 1--6 cover the detectable range in six log-spaced bands.
The bin edges are set as fixed multiples of the noise-floor standard
deviation $\sigma_n = 10^{-3}$:

\begin{equation}
  e_k = m_k \cdot \sigma_n, \quad
  \mathbf{m} = [3,\; 6,\; 15,\; 45,\; 150,\; 600],
  \label{eq:bin_edges}
\end{equation}

giving absolute edges $[0.003,\; 0.006,\; 0.015,\; 0.045,\; 0.15,\;
0.6]$.  The multipliers span roughly two orders of magnitude above the
noise floor and are calibrated so that each bin is occupied during typical
plume encounters - avoiding degenerate representations in which most
observations collapse to a single level.  A reading $c$ is assigned to bin
$b_k$ such that $e_{k-1} < c \leq e_k$ (bin 0 if $c \leq e_0$, bin 6 if
$c > e_5$).  Both antennae are quantized independently, yielding two
indices $b_L, b_R \in \{0, \ldots, 6\}$.

\paragraph{Wind-direction quantization.}

The wind direction is expressed relative to the agent's current heading
and quantized into eight equal octants of $45^\circ$ each.
Octant 0 corresponds to a headwind (wind arriving from directly ahead,
the most informative direction for upwind surge); octants increase
clockwise.  When wind speed falls below the calm threshold
($0.05\,\text{m/s}$), the direction is treated as octant 0 (headwind
default), which is the least informative assignment and penalizes neither
upwind nor downwind strategies.

\paragraph{Vocabulary summary.}

The full token vocabulary for one time step comprises:

\begin{center}
\begin{tabular}{llr}
\toprule
\textbf{Stream} & \textbf{Tokens} & \textbf{Size} \\
\midrule
Left concentration  & \textsc{blank}, C0--C5 & 7 \\
Right concentration & \textsc{blank}, C0--C5 & 7 \\
Wind direction      & octants W0--W7         & 8 \\
Action              & fwd, $\pm15^\circ$, $180^\circ$, cast$\pm$ & 6 \\
Special             & \textsc{pad}, \textsc{bos}, \textsc{eos}, \textsc{reset} & 4 \\
\bottomrule
\end{tabular}
\end{center}

The observation state at each step is the tuple $(b_L, b_R, w) \in
\{0,\ldots,6\}^2 \times \{0,\ldots,7\}$, giving $7 \times 7 \times 8 =
392$ unique states.  For the RL Q-network, this tuple is one-hot encoded
into a $22$-dimensional binary vector ($7 + 7 + 8$) and fed directly as
input.  For the generative world model, each stream is a separate embedding
lookup whose outputs are summed element-wise before entering the transformer,
following the multi-stream factored embedding design described below.

\subsection{Token-Based Generative World Model Interface}
\label{sec:world_model_interface}

The tokenizer is designed to serve a dual role: it provides the state
representation for the model-free RL agent used in the present work, and
it defines the token alphabet for a generative world model intended for
future deployment.  This section describes how the simulator trajectory
format maps to the sequence structure expected by the world model.

\paragraph{Action-conditioned sequence layout.}

At each control step the agent emits an action token and the environment
returns an observation token tuple.  A complete episode is serialized as
an interleaved observation--action sequence:
\begin{equation}
  \textsc{bos},\; \mathbf{o}_{0},\; a_{0},\; \mathbf{o}_{1},\; a_{1},\;
  \ldots,\; \mathbf{o}_{T-1},\; a_{T-1},\; \mathbf{o}_{T},\; \textsc{eos},
  \label{eq:sequence_layout}
\end{equation}
where $\mathbf{o}_t = (b_{L,t}, b_{R,t}, w_t)$ is the tokenized
observation at step $t$ and $a_t \in \{0,\ldots,5\}$ is the action index.
Each position in the sequence therefore carries a fixed grammatical
role—observation or action—that the model can learn to exploit.  The
\textsc{bos} and \textsc{eos} markers delimit episodes, and a
\textsc{reset} token can signal mid-sequence environment resets without
breaking the input stream.

\paragraph{Multi-stream factored embeddings.}

A key property of the vocabulary is that each time step's observation
decomposes into three \emph{independent} streams, each with its own
embedding matrix.  The embedding for step $t$ is formed by summing the
three per-stream lookup vectors:
\begin{equation}
  \mathbf{e}_t = E_L[b_{L,t}] + E_R[b_{R,t}] + E_W[w_t],
  \label{eq:factored_embedding}
\end{equation}
where $E_L, E_R \in \mathbb{R}^{7 \times d}$ and $E_W \in \mathbb{R}^{8
\times d}$ are learned embedding matrices.  Action embeddings are added
analogously at action positions.  This factored design allows the model to
share structure across antenna readings that differ only in wind direction,
and vice versa, without enumerating all $392$ joint observation types as
separate tokens—keeping the vocabulary size small enough for efficient
embedding lookup on resource-constrained hardware.

\paragraph{Training objective.}

The generative world model is trained by next-token prediction on
simulator-generated trajectories: given the sequence up to position $t$,
the model predicts the distribution over the next token.  At observation
positions this means predicting $P(\mathbf{o}_{t+1} \mid
\mathbf{o}_{\leq t}, a_{\leq t})$; at action positions, the model is
conditioned but does not predict (actions are supplied by the policy).
Because observations are factored into three streams, the prediction at
each observation position is decomposed into three independent softmax
outputs over 7, 7, and 8 classes respectively, giving a total cross-entropy
loss of
\begin{equation}
  \mathcal{L} = -\sum_{t} \Bigl[
    \log P(b_{L,t+1}) + \log P(b_{R,t+1}) + \log P(w_{t+1})
  \Bigr].
  \label{eq:world_model_loss}
\end{equation}
This supervision signal is dense - every step of every simulator trajectory
contributes three classification targets - while remaining entirely
derived from the physics simulator, with no human labeling or LLM
generation involved.

\paragraph{Role in planning.}

At inference time the trained world model replaces the physics simulator as an imagination engine.  
Given the agent's recent token history, the world model samples short rollouts of length $H$ for each candidate action, producing approximate future observation sequences without running the full filament simulation. 
A scoring function evaluates each rollout (e.g., expected whiff count, estimated distance reduction) and the action with the highest score is executed.  
The tokenizer's compact vocabulary - only 22 observation dimensions, 392 unique states - means the world model can be implemented as a small decoder-only transformer with a few hundred thousand parameters, meeting the real-time inference budget of most resource-constrained hardware.

%% file: src/supplementary/_experiment_hyperparams.tex
\section{Plume Navigation Hyperparameters}
\label{sec:experiment_hyperparams}

\begin{table}[h]
\centering
\caption{Hyperparameters for the plume navigation GPF agent.}
\label{tab:gpf_hyperparams}
\begin{tabular}{lll}
\toprule
\textbf{Parameter} & \textbf{Symbol} & \textbf{Value} \\
\midrule
\multicolumn{3}{l}{\textit{Environment}} \\
Domain size           & ---                    & $20 \times 20\,\text{m}$ \\
Time step             & $\Delta t$             & $0.1\,\text{s}$ \\
Episode budget        & $T_{\max}$             & $1200$ steps ($100\,\text{s}$) \\
Source emission rate  & $\lambda$              & $5\,\text{filaments/s}$ \\
Success radius        & $r_s$                  & $0.5\,\text{m}$ \\
Mean wind speed       & $\bar{v}$              & $1.0\,\text{m/s}$ \\
Wind correlation time & $\tau_c$               & $2.0\,\text{s}$ \\
\midrule
\multicolumn{3}{l}{\textit{Agent / network}} \\
Hidden width          & $d_h$                  & $64$ \\
Max hidden layers     & $L_{\max}$             & $4$ \\
Learning rate         & $\eta$                 & $10^{-3}$ \\
Discount factor       & $\gamma$               & $0.99$ \\
Initial exploration   & $\varepsilon_0$        & $1.0$ \\
Final exploration     & $\varepsilon_\infty$   & $0.05$ \\
Exploration decay     & ---                    & $0.9995$ per episode \\
\midrule
\bottomrule
\end{tabular}
\end{table}

\begin{table}[h]
    \caption{GPF Network Hyperparameters for Plume Task}
    \begin{center}
    \begin{tabular}{|c|c|c|}
    \hline
    \textbf{Value}& \textbf{Symbol}& \textbf{Description}\\
    \hline
    4500&$k_{\textbf{max}}$& Maximum training epochs\\
    64&$d_i$& $i^{th}$ layer dimension\\
    0&$\rho_e$& Epoch patience required before GPF is enabled\\
    1000&$\rho_k$& Epoch patience required before growing a layer\\
    500&$\rho_p$& Epoch patience required before pruning neurons\\
    3000&$\rho_f$& Epoch patience required before a neuron can be frozen\\
    $1e-3$&$\omega_v$& Loss stagnation threshold for layer growth\\
    $1e-6$&$\omega_p$& Belief threshold for pruning\\
    0.01&$\omega_f$& Weight-change threshold for layer freezing\\
    $1e-3$&$\alpha$& Learning rate\\
    0.99&$\gamma$& Discount factor\\
    0.05&$\epsilon$& Exploration greediness\\
    0.9995&$\eta_g$& Greediness decay\\
    0&$\eta_{bs}$& Batch size decay\\
    1&$\eta_h$& Belief value increment\\
    \hline
    \end{tabular}
    \label{tab:paramsTable}
    \end{center}
\end{table}

%% file: src/supplementary/_results_table.tex
\section{Episode-by-Episode Training Results}
\label{sec:training_curve_table}

\begin{table}[h]
\centering
\caption{Evaluation success rate and network state throughout training. 4500 episodes were established as the limit in order to timebox experiments. (G = grow event; P = prune event)}
\label{tab:gpf_training_curve}
\begin{tabular}{llll}
\toprule
\textbf{Episode} & \textbf{Success (\%)} & \textbf{Layers} & \textbf{Event} \\
\midrule
500  & 24.5          & 1 & --- \\
1000 & 95.5          & 1 & --- \\
1500 & 40.5          & 1 & --- \\
2000 & 97.5          & 2 & G: layer $1 \to 2$ \\
2000 & ---           & 2 & P: kept $47.9\%$ \\
2500 & 91.5          & 2 & --- \\
3000 & \textbf{98.0} & 3 & G: layer $2 \to 3$ (best checkpoint) \\
3000 & ---           & 3 & P: kept $19.6\%$ (2 layers) \\
3500 & 95.5          & 3 & --- \\
4000 & 96.5          & 4 & G: layer $3 \to 4$ \\
4000 & ---           & 4 & P: kept $9.9\%$ (3 layers) \\
4500 & 95.5          & 4 & --- \\
\bottomrule
\end{tabular}
\end{table}

%% file: src/supplementary/_experiments_b.tex
\section{GPF Generalization}
\label{sec:gpf_general_experiments}
To demonstrate the \textit{grow-prune-freeze} methodology, we selected three different tasks with many known benchmarks in literature: image classification over CIFAR10, reinforcement learning over the Atari environment, and text generation via GPT-2.

For each task, we trained a baseline model and a GPF variant of the same model.
All models were trained under identical parameters with only one exception: the baseline model begins training at \textit{n} hidden layers while the GPF model begins training with one hidden layer and grows to a maximum depth of \textit{n} layers.
All hyperparameters, data, training procedures, and seeds are identical otherwise.
The hyperparameters for each model can be observed in the following sections.
GPF-specific hyperparameters can be found below in Table \ref{tab:gpf_hyperparams}.

\begin{table}[h]
    \caption{GPF Classifier Hyperparameters}
    \begin{center}
    \begin{tabular}{|c|c|c|}
    \hline
    \textbf{Method}& \textbf{Description}\\
    \hline
    $\rho_e$& Epoch patience required before GPF is enabled& 7\\
    $\rho_k$& Epoch patience required before growing a layer& 5\\
    $\rho_p$& Epoch patience required before pruning neurons& 2\\
    $\rho_f$& Epoch patience required before a neuron can be frozen& 7\\
    $\omega_v$& Weight threshold for belief counter& 1e-2\\
    $\omega_p$& Activation threshold for pruning& 0.5\\
    $\omega_f$& Loss threshold for neuron freezing& 1e-1\\
    $\gamma$& Discount factor (RL) &0.99\\
    $\epsilon$& Greediness (RL) &1.0\\
    $\eta_g$& Greediness decay (RL) &0.995\\
    $\eta_{bs}$& Batch size decay&0.5\\
    $\eta_h$& Belief value decay&0.1\\
    $\beta_w$& Weight threshold for belief increment&0.1\\
    $\beta_act$& Post-activations above this increase belief counter&0.1\\
    \hline
    \end{tabular}
    \label{tab:gpf_hyperparams}
    \end{center}
\end{table}

\subsection{GPF Task: CIFAR Image Classification}
\begin{table}[h]
    \caption{GPF Classifier}
    \begin{center}
    \begin{tabular}{|c|c|}
    \hline
    \textbf{Parameter}& \textbf{Value}\\
    \hline
    layers& 4\\
    layer types& Fully connected only\\ 
    input dimension& 3072\\
    output dimension& 10\\
    hidden dimension& 512\\
    batch size& 128\\
    learning rate& 1e-4\\
    activation functions& ReLU\\
    epochs& 30\\
    loss function& Cross Entropy\\
    optimizer& Adam\\
    pruning technique& LeCun\\
    \hline
    \end{tabular}
    \label{tab:sm_gpf_cifar}
    \end{center}
\end{table}

\begin{figure}[h]
  \centering
  \includegraphics[width=\linewidth]{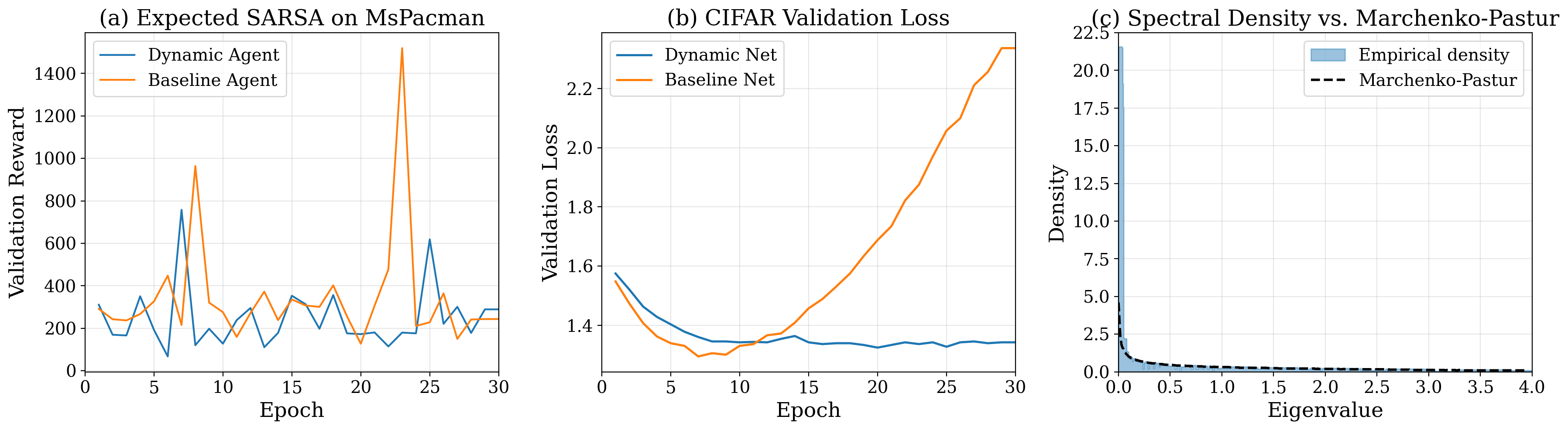}
  \caption{(a) Validation losses of a GPF policy and a baseline static policy trained under identical parameters for the Atari task. Both policies are neural networks based on Expected SARSA. (b) Validation losses of a GPF network and a baseline static neural network trained under identical parameters for the CIFAR image classification task. (c) The eigenvalue spectra for the weights of the GPF network from the CIFAR image recognition task follow the Marchenko-Pastur distribution, maintaining the hypothesis of \cite{Pennington2017}.}
  \label{fig:gpf_general_comparison}
\end{figure}

We selected image classification over CIFAR10 because it provides a foundational entry point and because it allows us to build off the assumptions and experiments from the similar work by Dohare, et al. \cite{Dohare2024_plasticity_loss}.
In reference to Figure \ref{fig:gpf_general_comparison}, one will notice that the validation loss begins to diverge for the baseline model, but steadily decreases for the dynamic GPF model. 
This may indicate that GPFs can provide more stable learning even while simultaneously adding layers and pruning neurons.

Figure \ref{fig:gpf_general_comparison} shows the weight eigenvalues after each layer is added.
In observation of the eigenvalue spectra over several different CIFAR batches, the empirical spectral density (ESD) of the eigenvalues indeed follow the Marchenko-Pastur distribution - a characteristic that Pennington, et al. and Martin, et al. indicate provides evidence of stable weight eigenvalues during learning \cite{Pennington2017} \cite{Martin2021RMT}.


\subsection{GPF Task: Atari Reinforcement Learning}
\begin{table}[h]
    \caption{GPF Agent}
    \begin{center}
    \begin{tabular}{|c|c|}
    \hline
    \textbf{Parameter}& \textbf{Value}\\
    \hline
    layers& 4\\
    layer types& Fully connected only\\ 
    input dimension& 3072\\
    output dimension& 10\\
    hidden size& 128\\
    batch size& 16\\
    learning rate& 1e-3\\
    activation functions& ReLU\\
    epochs& 30\\
    max steps& 150\\
    loss function& MSE\\
    optimizer& Adam\\
    pruning technique& LeCun\\
    policy& Expected SARSA\\
    \hline
    \end{tabular}
    \label{tab:sm_gpf_atari}
    \end{center}
\end{table}

Reinforcement learning contains many tasks and benchmarks from which one could select, but we evaluated over the Atari environment in order to assess GPF's decision making capabilities.
Specifically, we select the \textit{Ms Pacman} environment as it is seemed an appropriate navigation analog to our olfactory tasks.
Conclusions are difficult to draw from the accompanying figure in \ref{fig:gpf_general_comparison}, but one can immediately observe that the volatility in validation rewards for the baseline policy is much higher than that of the dynamic GPF policy.


A more comprehensive evaluation over the entire Atari suite is an intuitive segue for future work.

\subsection{GPF Task: GPT-2 Language Modeling}
\FloatBarrier
\begin{table}
    \caption{GPF Autoregressive Model}
    \begin{center}
    \begin{tabular}{|c|c|}
    \hline
    \textbf{Parameter}& \textbf{Value}\\
    \hline
    layers& 12\\
    layer types& GPT Block\\ 
    vocab size& 50257\\
    sequence length& 64\\
    block size& 32\\
    attn heads/block& 8\\
    attn mechanism& Causal Self-Attention\\
    embedding dimension& 64\\
    batch size& 8\\
    learning rate& 1e-4\\
    activation functions& ReLU, GeLU\\
    epochs& 10000\\
    loss function& Cross Entropy\\
    optimizer& Adam\\
    pruning technique& LeCun\\
    \hline
    \end{tabular}
    \label{tab:sm_gpf_gpt2}
    \end{center}
\end{table}

Finally, given today's attraction toward and the many recent advances in large language models, we select GPT-2 as a means to show how GPFs can be extended to generative tasks and much larger architectures.
We acknowledge that GPT-2 is no longer considered ``state-of-the-art'', but the thorough evaluation of GPFs over the much larger state-of-the-art architectures is outside of the financial and computational budget for a single PhD student.
The architectures of many of the notable large language models today were derived from GPT-2 \cite{sebastianraschkaFromGPT2, olmo2025olmo3, openai2025gptoss120bgptoss20bmodel}, so we intuit that using it as a vehicle for analysis is a representative architecture of what might occur in large model.

\begin{figure}
  \centering
  \includegraphics[width=\linewidth]{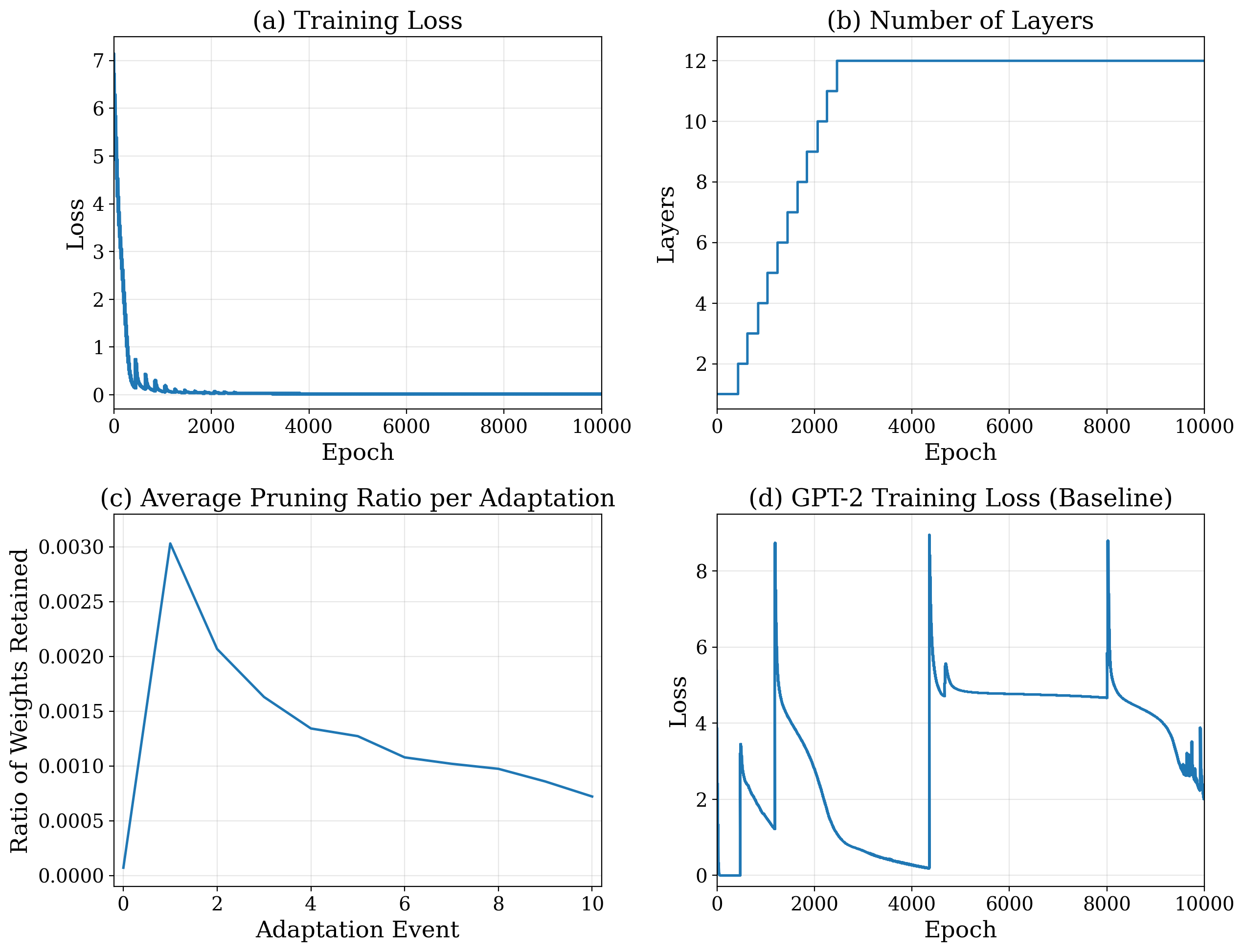}
  \caption{(a) Training loss of the GPT-2 baseline; training parameters were identical to the GPF variant below. (b) The validation loss shows steady convergence over training, even while growing and pruning its own parameters. (c) The GPF reached the maximum number of layers at around 2500 epochs and kept training. (d) The pruning ratio over each adaptation event shows that the number of weights pruned is much higher at the beginning of training, but tapers as eigenvalues stabilize.}
  \label{fig:gpf_gpt2_baseline}
\end{figure}

We use the \textit{TinyShakespeare} dataset from Karpathy \cite{karpathy2015_tinyshakespeare} to train a baseline model and GPF variant of GPT-2.
Figure \ref{fig:gpf_gpt2_baseline} shows the final loss plot for the baseline GPT-2 model.
Interestingly, one can see that convergence is not achieved and the loss is very volatile.
Conversely, the training loss for the GPF variant of GPT-2 in Figure \ref{fig:gpf_gpt2_baseline} shows good convergence, with a steady decay after epoch 2000.
GPF GPT-2 achieved the maximum layer count at around epoch 2500 and ceased growing for the rest of training.
Consistent with the work of LeCun \cite{lecun1989optimalbraindamage} and Frankle, et al. \cite{frankle2019lotterytickethypothesisfinding}, the percentage of neurons pruned decays as training progresses.
For GPFs specifically, this may indicate that neurons from earlier weights have hardened their belief states around values that reflect true patterns in the data.

Our intent with the tasks above was to show initial evidence that the principles of the GPF framework can generalize to different models and tasks.
There is much more to investigate in relation to the Pareto frontiers of the conventional ``train-and-freeze'' paradigm -  just as humans continuously learn and update priors as they gather more experiences, so can we enable AI models to do the same.
However, we acknowledge that our work here is only evidence supporting the GPF framework for continual learning, and that a thorough evaluation over a large battery of tasks and different architecture types is prudent future work.



